\documentclass[runningheads]{llncs}
\usepackage{amsmath,amsfonts,amssymb}
\usepackage{subfig}
\usepackage{adjustbox}
\usepackage{makeidx}
\usepackage{float}
\usepackage{multirow}
\usepackage{subfig}
\usepackage{graphicx}
\graphicspath{ {./images/} }
\usepackage{booktabs}
\usepackage{xcolor}
\usepackage{url}            
\usepackage{booktabs}       
\usepackage{amsfonts}   
\usepackage{subfig}         
\usepackage{nicefrac}       
\usepackage{microtype}      
\usepackage{color}
\usepackage{multirow}
\usepackage[colorlinks=true, allcolors=blue]{hyperref}

\begin{document} 
\title{Zero Shot Learning for Multi-Modal Real Time Image Registration}

\author{Avinash Kori\orcidID{0000-0002-5878-3584} \and
Ganapathy Krishnamurthi\orcidID{0000-0002-9262-7569}}
\authorrunning{A.Kori et al.}
\institute{Department of Engineering Design, Indian Institute of Technology Madras, India\\
\email{gankrish@iitm.ac.in}}

\maketitle      

\begin{abstract}
We present an unsupervised image registration framework, using a pre-trained deep neural network as a feature extractor. We refer this to zero-shot learning, due to non overlap between training and testing dataset (none of the network modules in the processing pipeline were trained specifically for the task of medical image registration). Highlights of our technique are: (\textbf{a}) No requirement of a training dataset (\textbf{b}) Keypoints i.e. locations of important features are automatically estimated (\textbf{c}) The number of key points in this model are fixed and can possibly be tuned as a hyperparameter. (\textbf{d}) Uncertainty calculation of the proposed, transformation estimates (\textbf{e}) Real-time registration of images. Our technique was evaluated on BraTS, ALBERT and collaborative hospital Brain MRI data. Results suggest that the method proved to be robust for affine transformation models and the results are practically instantaneous, irrespective of the size of the input image.
\end{abstract}

\section{Introduction}
\label{sec:intro}
Image registration is one of the key image processing steps in medical image analysis. It involves spatial alignment of certain features from multiple sets of images. Real-time image registration plays a crucial role in image-guided surgery, in aligning pre-operative and intra-operative images. For instance, the opening of skull and dura during neurosurgical procedures leads to leakage of CSF which in-turn results in a change of brain shape. In case of tumor removal, the registration becomes very hard problem as all the features in intra-operative (fixed) images don't have correspondence in pre-operative (moving) image, due to the removal of tumor tissue. In this paper, we propose an efficient learning based real-time image registration method which doesn't require training data or priors.Traditionally, image registration is performed by exploiting the pixel intensity information between pairs of fixed and moving images. Many recent advancements in deep learning have enabled image registration to be performed using Convolutional Neural Network (CNN) \cite{DIRNet}. A  convolutional stacked autoencoder \cite{CAE} based approach was used to extract features from the fixed and moving images which were subsequently used in registration algorithms. However,  in all these methods, the task of registration was performed on the entire image rather than a few selected features, which makes this registration problem computationally very expensive and not suitable for real-time applications. In the proposed method, we use a pre-trained CNN for feature extraction followed by a key point detector. The key points are fed to the Multi-Layered Perceptron(MLP) based regression module so as to estimate the transformation parameters such as scale, rotation, shear, translation etc.  A typical registration framework would consist of a feature extractor, keypoint matching module followed by a parameter estimator. Neural network approaches typically tend to mimic the above sequence with individual layers (or sets of layers) to perform each of the aforementioned tasks. The parameters of the network are learned using the standard backpropagation algorithm. 
The rest of the paper follows as materials and methodology in section 2 and results and discussion in section 3. The paper ends with conclusions and future work in section 4.

\section{Materials and Methodology}
\begin{figure}[h]
\centering
\includegraphics[width=1\textwidth]{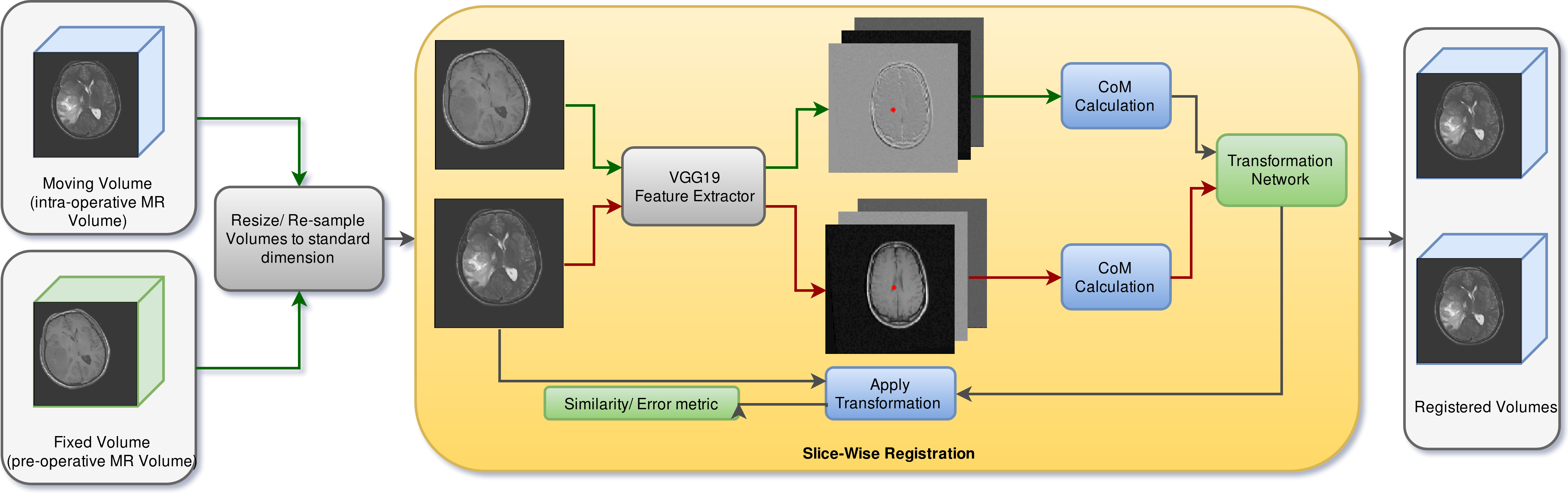}
\caption{Schematics of our method used for registration. Model takes pair of fixed and moving images as an input, the pre-trained Convolutional network provides feature map outputs. These feature maps are used in CoM calculation which in-turn is used in parameter estimation by pre-trained MLP.}
\label{direct_pipeline}
\end{figure}
\subsection{Data}
Our method was tested on different datasets of brain MR images taken from the ALBERT dataset \cite{IXI} for clinical image analysis (20 volumes with T1 and T2 sequences), BraTs \cite{brats1},\cite{brats2}, \cite{brats3}, \cite{brats4} dataset (240 volumes of 240x240x155 with T1, T1ce, T2 and FLAIR sequences) and finally on a dataset collected from a collaborative hospital with ethical clearance. 

\subsection{Preprocessing of data}
\par Very minimal pre-processing was performed on the fixed and the moving image before feeding into the pipeline. The images were mean centered and normalized followed by padding with zeros to prevent loss of information due to large rotations or translations.

\subsection{Feature Extractor}
The feature extractor in the pipeline aids in dynamically extracting important features from both moving and fixed images. These extracted features are subsequently used for estimating the affine matrix. We make use of a VGG-19 \cite{vgg} model pre-trained on natural images as the feature extractor. Generally, the initial layers help in capturing the scale and translational invariant features like edges \& corners, while the deeper layers help in localizing the object in an image, which aids in image classification. To utilize this property of CNN, we consider the feature maps from the first two layers of the network for Center of Mass (CoM) calculation and affine parameter estimation.

\par {As only primary features like edges, corners were used in CoM estimation in our proposed method, a 2-3 layered convolutional neural network (CNN) trained on relevant data might be sufficient for the task of feature extraction, but we made use of the pre-trained model with transfer learning approach, exploiting the features learned from a rich imagenet \cite{imagenet} dataset, In this case VGG serves to be the smallest and computationally inexpensive model to use.}
Thresholding of the feature maps, Eq. \ref{thres} was done to obtain more significant features from both the moving and fixed images. In our case, we used 95\%  of maximum activation (hyperparameter, experimentally determined) in the feature maps as the threshold. These high activation regions in the feature maps are further considered for CoM calculation.
\begin{equation}
F[F< = 0.95 \times max(F)] = 0
\label{thres}
\end{equation}
where F corresponds to a single feature map obtained from the network. Thresholding of images helps in capturing significant features from an image, which helps in the registration of pre, and intraoperative images during tumor resection.

\subsection{CoM Extraction}
Thresholded feature maps obtained from the feature extraction step were used for CoM calculation. Here CoM is a  centrality measure of feature maps. Mathematically CoM calculation is done based on equations \ref{comvec}:
\begin{equation}
    CoM_X  = \frac{\Sigma \Sigma f_{ij} x_{ij}}{ \Sigma \Sigma f_{ij}}; \quad  CoM_Y = \frac{\Sigma \Sigma f_{ij} y_{ij}}{ \Sigma \Sigma f_{ij}}
    \label{comvec}
\end{equation}
where, $f_{ij}$ is the pixel intensity at the i, jth position $x_{ij} \quad and \quad y_{ij}$ are the values of x and y at i$^{th}$, j$^{th}$ position respectively. The CoMs in the moving and fixed image corresponds to the key points in the moving and fixed image respectively. The key points were then normalized while maintaining the aspect ratio of the image. The normalized key points then input to the regressor network to estimate the transformation parameters.
\subsection{Transformation Estimator}
\begin{figure}[h]
\centering
\includegraphics[width=.5\textwidth]{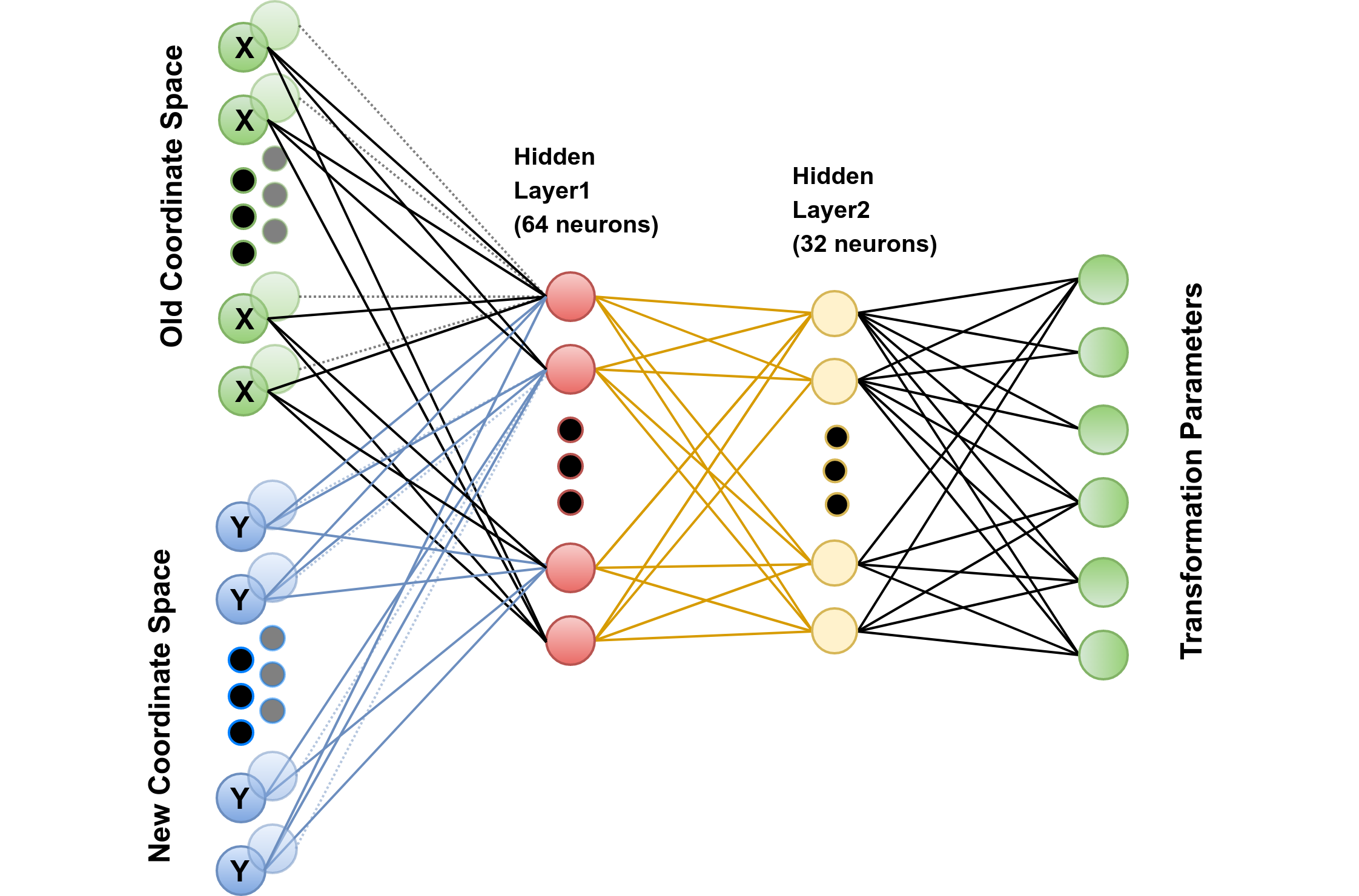}
\caption{Above figure shows architecture used in transformation network. Network takes data points from two different co-ordinate spaces and gives output as 6 Transformation parameters between both the coordinate spaces used. In case of above network the Rectified Linear Unit (ReLU) was used as an activation function. Final layer is just a linear layer (to allow negative values)}
\label{training_strategy}
\end{figure}
\subsubsection{Training Strategy}
The neural network to estimate the registration parameters was trained by generating the a set of random data points say $X$ (old co-ordinates space) from a uniform distribution between 0 and 1.0. A random transformation matrix, $M$, was generated from a uniform distribution between -1.0 and 1.0. The new coordinate space, Y, was generated by applying the random transformation on to the old co-ordinates space such that Y = MX. The network was trained to estimate the transformation matrix, M, given $X , Y$. The trained network thus does not depend on the input image data format but rather on key-point locations. This strategy enables us to train transformer networks independently for different type of transforms. In this paper, we have only considered affine transform.

\subsubsection{Parameter Estimation}
\par In the context of image registration, the keypoints detected from the fixed and moving image forms the data points for the old coordinate space ($X$) \& new coordinate space ($Y$) respectively. The transformation network yields the transformation metric, $M$, given the $X, Y$. A total of 128 keypoints were extracted from both the fixed and moving images. Ten sets of old and new coordinate space were generated by randomly sampling  64 keypoints with replacement from extracted keypoints. For each set of coordinate space data, the trained transformation network predicts the appropriate transformation parameters. The final transformation applied on to the moving image was equivalent to the average of the transformation parameters on the ten sets. This approach was done so as to reduce the effect of noise in the transformation.

\subsection{Similarity/Error Metric}
{We have verified our model performance using various similarity and error metrics, it turns out to be that our model performs better when evaluated with any of the metric below.}

\subsubsection{Dice score}
Dice score is a metric most commonly used in image segmentation used to find an extent of overlap between two images. Here we estimate dice to find an overlap between moving and fixed image. Dice score is given by eqn. \ref{dice}.
\begin{equation}
\label{dice}
Dice Score = \frac{2\times FixedImage \cap MovingImage}{FixedImage \cup MovingImage}
\end{equation}
\subsubsection{Structural Similarity}
Structural similarity index (SSIM) is measure of similarity between two images. SSIM considers local pixel information for score calculation. This means it carries an idea of spatial positioning of pixels in an images. SSIM if given by eqn. \ref{SSIM}.
\begin{equation}
\label{SSIM}
{\hbox{SSIM}}(F, M)={\frac  {(2\mu _{F}\mu _{M}+c_{1})(2\sigma _{{FM}}+c_{2})}{(\mu _{F}^{2}+\mu _{M}^{2}+c_{1})(\sigma _{F}^{2}+\sigma _{M}^{2}+c_{2})}}
\end{equation}
In the above equation F corresponds to Fixed image and M corresponds to Moving image. $c_1, c_2$ are two variables to stabilize week denominators. SSIM $\in$ [0, 1.0], where 0 being least and 1 being maximum score.
\subsubsection{Mutual Information}
{Mutual information (MI) is one of the quantities which measures the amount of correlation between two different random variables. In case of registration as higher the MI score as good the performed registration.} MI between two variables is given by eqn. \ref{MI}.
\begin{equation}
\label{MI}
MI(F, M) = \mathbf{E}(P_{FM}(F,M)) \times log(\frac{P_{FM}(F,M)}{P_{F}(F)P_{M}(M)})
\end{equation}
where $P_{FM}(F, M)$ denotes joint distribution, $\mathbf{E}$ denotes expectation value and $P_F(F), P_M(M)$ denotes marginals. MI $\in$ [0, 1.0], where 0 being least and 1 being maximum score
\subsubsection{Mean Squared Error}
Mean Squared Error (MSE) is an error metric which finds the pixel wise deviation between fixed and moving image. MSE is given by equation \ref{MSE}.
\begin{equation}
\label{MSE}
MSE = \frac{1}{H\times W}\sum_{x =0}^{W} \sum_{y=0}^{H} (FixedImage_{x,y} - MovingImage_{x,y})^2
\end{equation}
where $FixedImage_{x,y}$ denotes pixel value at x,y position in the Fixed Image, similarly for $MovingImage_{x,y}$ denotes pixel value at x,y position in the moving image. MSE $\in$ [0, $\infty$), as lesser the MSE as better the registration.

\subsection{Uncertainty Estimation}
 Due to the unavailability of ground truths in a real-time image registration system, an uncertainty estimate would help gauge system reliability. Uncertainty in the transform parameters is estimated by random blackening of pixels in a moving image and estimating transformation parameters with respect to fixed image \cite{uncertaintyIR}. This experiment was repeated (n = 10) and the variance between the estimated parameters was considered as the uncertainty associated with the given fixed and moving image. However, the neural network transform estimator enables one to perform the boot strap uncertainty estimation in real time.
 
 \subsection{Inference}
 \par To test the robustness of the proposed technique, apart from the inherent deformation between the fixed \& moving image, we additionally induce artificial transformations on the moving image before registration. These artificial transformation included translational upto 50 pixels in both $x$ \& $y$ direction, rotation upto 0.3 radians (17.18 degrees) \& a shear upto 0.03 radians (1.71 degrees). Model was evaluated on various brain MR datasets and also on natural image datasets. Registration score was gauged using dice, structural similarity, mutual information and mean squared error as the metrics, results of each metric on each of these datasets are illustrated in section below.
 During inference both moving and fixed images were passed through VGG-19, CoM from the extracted features were estimated. CoMs, as a centrality measure of a feature, was used by the regressor module to estimate the transformation parameters. In the scenario where the moving data is highly skewed when compared to the fixed image, the transformation parameter could be estimated in a iterative fashion with a pre-defined learning rate. 

\subsection{Comparison}
\par To compare and benchmark our results, we considered two standard nearly realtime registration techniques SIFT(scale invariant feature transformation) features based image registration and SimpleElastix based image registration. The performance was compared on the bases of MI, SSIM, Dice, RMSE and time. The results are tabulated below in the results section (table \ref{comparison_table}).



\section{Results and Discussion}
\begin{figure}[h]
\centering
\subfloat[Fixed (T2)]{\includegraphics[width=0.19\textwidth,keepaspectratio]{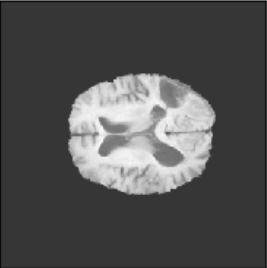}}\hfill
\subfloat[Moving (T2)]{\includegraphics[width=0.19\textwidth,keepaspectratio]{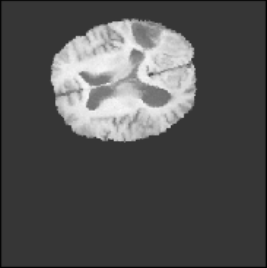}}\hfill
\subfloat[Before]{\includegraphics[width=0.19\textwidth,keepaspectratio]{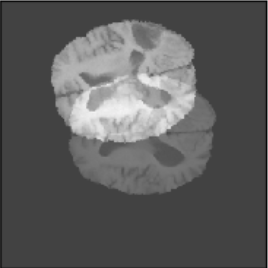}}\hfill
\subfloat[After]{\includegraphics[width=0.19\textwidth,keepaspectratio]{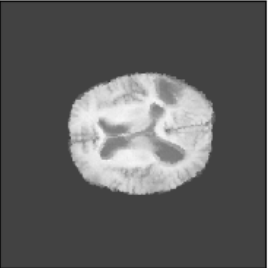}}\hfill
\subfloat[Uncertainty]{\includegraphics[width=0.19\textwidth,keepaspectratio]{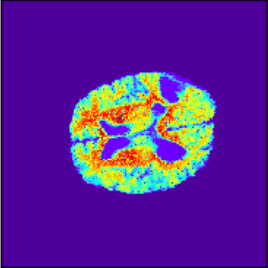}}\\
\subfloat[Fixed (T1)]{\includegraphics[width=0.19\textwidth,keepaspectratio]{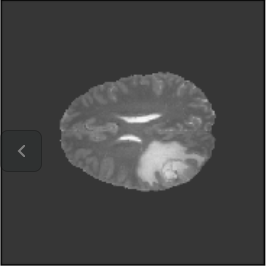}}\hfill
\subfloat[Moving (T2)]{\includegraphics[width=0.19\textwidth,keepaspectratio]{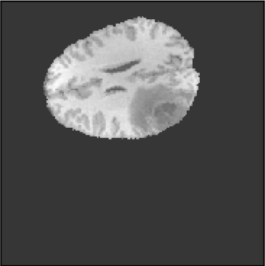}}\hfill
\subfloat[Before]{\includegraphics[width=0.19\textwidth,keepaspectratio]{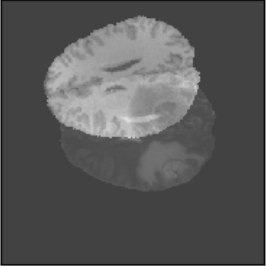}}\hfill
\subfloat[After]{\includegraphics[width=0.19\textwidth,keepaspectratio]{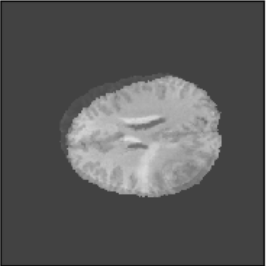}}\hfill
\subfloat[Uncertainty]{\includegraphics[width=0.19\textwidth,keepaspectratio]{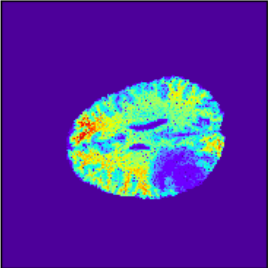}}
\caption{Intra and Inter modality MR registration. Time taken: 0.21 sec}
\label{unimodalbrain}
\end{figure}

\begin{figure}[h]
\subfloat[]{\includegraphics[width=0.25\textwidth,keepaspectratio]{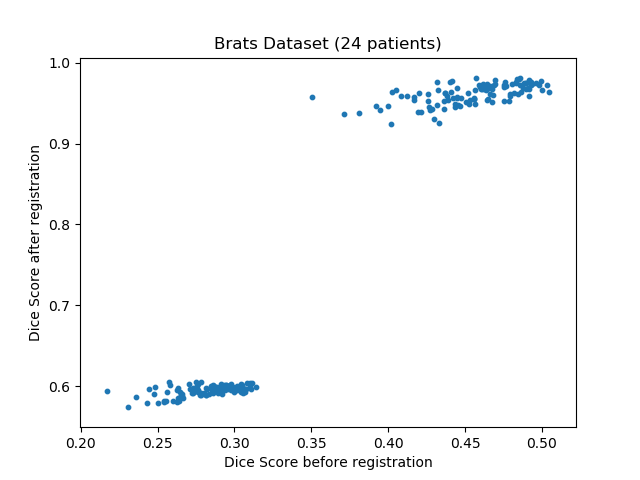}}\hfill
\subfloat[]{\includegraphics[width=0.25\textwidth,keepaspectratio]{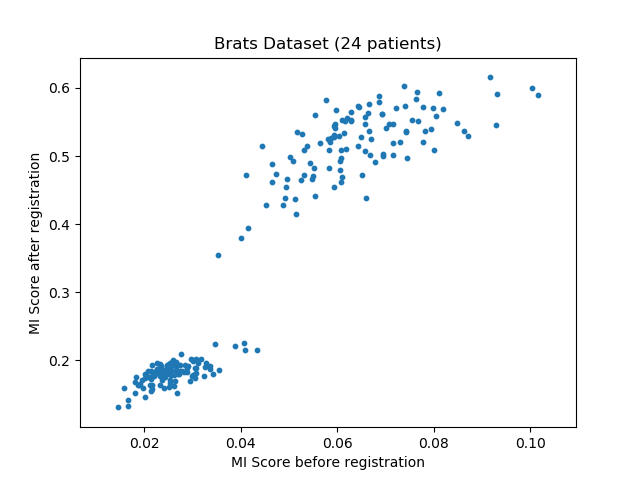}}\hfill
\subfloat[]{\includegraphics[width=0.25\textwidth,keepaspectratio]{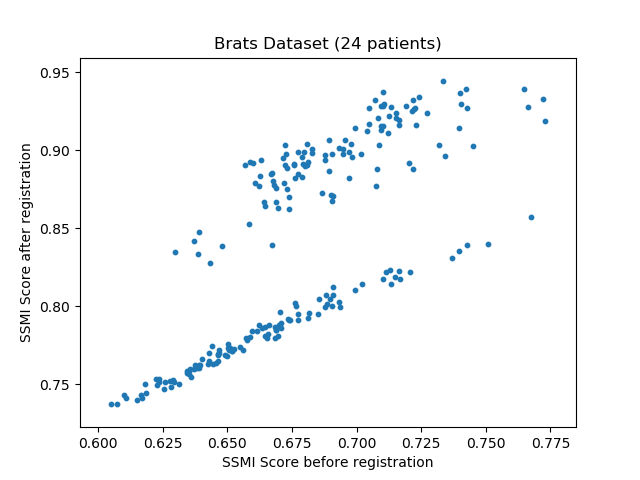}}\hfill
\subfloat[]{\includegraphics[width=0.25\textwidth,keepaspectratio]{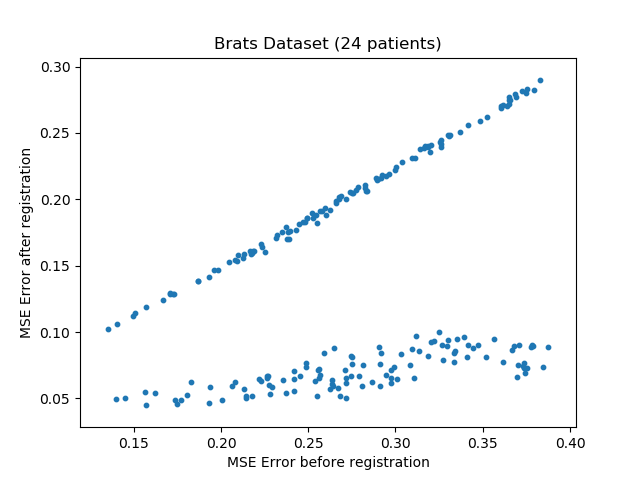}}\hfill
\caption{Scatter plots between similarity/ error score obtained before and after registration on the BraTs data. sub figure (a) illustrates the effect of registration on Dice score (x $\in$ [0.20, 0.50], y $\in$ [0.60, 1.0]), (b) illustrates the effect of registration on MI score (x $\in$ [0.02, 0.10], y $\in$ [0.20, 0.60]), (c) illustrates the effect of registration on SSMI score (x $\in$ [0.60, 0.77], y $\in$ [0.75, 0.95]) and (d) illustrates the effect of registration on MSE error (x $\in$ [0.15, 0.40], y $\in$ [0.05, 0.30])}
\label{fig:scatter_images_brats}
\end{figure}
To illustrate the performance of the proposed technique for clinical applications, MR images were chosen. The performance of the technique was tested by registering multi-modal MR image pairs. In a clinical setup, it is common practice to register images or volumes acquired from different imaging modalities or MR sequences. Figure  \ref{unimodalbrain} illustrates the degree of overlap between moving images (a) T2 and (f) T1 weighted sequence on fixed images (b) T2 and (g) T2 weighted sequence respectively. It was observed using the registration proposed in this manuscript, the dice score improved from 0.44 (prior to registration) to 0.91 (post registration) for multimodal sequences. On the BraTs data,  figure \ref{fig:scatter_images_brats} (a-d) illustrates the performance of our method based on other metrics such as Dice, SSMI, MI, and MSE respectively. As the key points were generated by utilizing only the feature maps of the initial two layer of the network, the spatial dimension of the data has little impact on the performance. Unlike other CNN based technique which considers the entire image for estimating the transformation parameters, our method requires CoMs which reduces the dimensionality from (m $\times$ n $\times$ n) to (m $\times$ 2), where m denotes the number of feature maps and n is the pixel dimension of the feature map. This aids in lowering execution time drastically. The proposed pipeline also consists of an uncertainty estimator. uncertainty maps illustrated in all the above examples show the region of higher variance in registration, which in turn suggests surgeons/ observers to make better decisions. Pixel region with red color indicates the larger variance region while blue indicated low variance region. For example in figure \ref{unimodalbrain} (e) shows that the central region of the brain has higher pixel-wise variance as compared to the outer part of the brain. The performance of the algorithm on volumes on the same modality is illustrated in figure \ref{unimodalbrain} (a-e)

\begin{table}[]
\centering
\caption{This table shows the comparison between various techniques used for image registration in terms of Dice, MI, SSIM, MSE and time taken. }
\label{comparison_table}
\begin{tabular}{|l|l|l|l|l|l|}
\hline
\multirow{2}{*}{Techniques} & \multicolumn{4}{l|}{Change (after - before) registration}             & \multirow{2}{*}{\begin{tabular}[c]{@{}l@{}}Time take per \\ image in sec.\end{tabular}} \\ \cline{2-5}
                            & Dice            & MI              & SSIM            & MSE             &                                                                                            \\ \hline
SimpleElastix               & + 0.263         & +0.221          & +0.404          & -0.0361         & 1.6                                                                                        \\ \hline
SIFT Based                  & +0.248          & +0.15           & +0.383          & -0.041          & 0.3                                                                                        \\ \hline
\textbf{Ours}               & \textbf{+0.294} & \textbf{+0.373} & \textbf{+0.431} & \textbf{-0.072} & \textbf{0.2}                                                                               \\ \hline
\end{tabular}
\end{table}

\section{Conclusion and Future work}
We present an unsupervised deep learning based method for image registration. A pre-trained VGG-19 network was used as the feature extractor, while COMs serve as the measure of centrality of a features. Based on the COMs of the moving \& fixed image, the regressor module estimates the parameters constituting the transformation matrix.  The neural networks namely VGG-19 \& transformation network used in this manuscript were trained on non-medical datasets, it makes the training and testing data completely disjoint. Due to the reason stated above, the proposed technique falls under the category of "Zero-shot learning". The robustness of the technique was estimated by testing the technique on a variety of medical and natural image database. Based on the preliminary results, we conclude, that \textbf{(a)}The proposed technique could be applied to learn any generic transform from no or little availability of data. \textbf{(b)}The capability of our technique is not limited to medical images, but it could be extended to the natural image provided perspective transforms between images are minimal. In future we plan to extend this work on natural deformations insted of artificially inducing them, followed by 3D volumetric registration.

\bibliography{report} 

\begin{thebibliography}{10}

\bibitem{DIRNet}
B.~D. de~Vos, F.~F. Berendsen, M.~A. Viergever, M.~Staring, and I.~I{\v{s}}gum,
  ``End-to-end unsupervised deformable image registration with a convolutional
  neural network,'' in {\em Deep Learning in Medical Image Analysis and
  Multimodal Learning for Clinical Decision Support},  pp.~204--212, Springer,
  2017.

\bibitem{CAE}
J.~Masci, U.~Meier, D.~Cire{\c{s}}an, and J.~Schmidhuber, ``Stacked
  convolutional auto-encoders for hierarchical feature extraction,'' in {\em
  International Conference on Artificial Neural Networks},  pp.~52--59,
  Springer, 2011.

\bibitem{IXI}
``Ixi - information extraction from images available at:
  http://www.brain-development.org/ [accessed 08 april. 2018]..''

\bibitem{brats1}
B.~H. Menze, A.~Jakab, S.~Bauer, J.~Kalpathy-Cramer, K.~Farahani, J.~Kirby,
  Y.~Burren, N.~Porz, J.~Slotboom, R.~Wiest, {\em et~al.}, ``The multimodal
  brain tumor image segmentation benchmark (brats),'' {\em IEEE transactions on
  medical imaging}~{\bf 34}(10), pp.~1993--2024, 2015.

\bibitem{brats2}
S.~Bakas, H.~Akbari, A.~Sotiras, M.~Bilello, M.~Rozycki, J.~S. Kirby, J.~B.
  Freymann, K.~Farahani, and C.~Davatzikos, ``Advancing the cancer genome atlas
  glioma mri collections with expert segmentation labels and radiomic
  features,'' {\em Scientific data}~{\bf 4}, p.~170117, 2017.

\bibitem{brats3}
S.~Bakas, H.~Akbari, A.~Sotiras, M.~Bilello, M.~Rozycki, J.~S. Kirby, J.~B.
  Freymann, K.~Farahani, and C.~Davatzikos, ``Advancing the cancer genome atlas
  glioma mri collections with expert segmentation labels and radiomic
  features,'' {\em Scientific data}~{\bf 4}, p.~170117, 2017.

\bibitem{brats4}
S.~Bakas, H.~Akbari, A.~Sotiras, M.~Bilello, M.~Rozycki, J.~Kirby, J.~Freymann,
  K.~Farahani, and C.~Davatzikos, ``Segmentation labels and radiomic features
  for the pre-operative scans of the tcga-lgg collection,'' {\em The Cancer
  Imaging Archive} , 2017.

\bibitem{vgg}
K.~Simonyan and A.~Zisserman, ``Very deep convolutional networks for
  large-scale image recognition,'' {\em arXiv preprint arXiv:1409.1556} , 2014.

\bibitem{imagenet}
J.~Deng, W.~Dong, R.~Socher, L.-J. Li, K.~Li, and L.~Fei-Fei, ``{ImageNet: A
  Large-Scale Hierarchical Image Database},'' in {\em CVPR09},  2009.

\bibitem{uncertaintyIR}
J.~Kybic, ``Bootstrap resampling for image registration uncertainty estimation
  without ground truth,'' {\em IEEE Transactions on Image Processing} , 2010.

\end{thebibliography}
\bibliographystyle{spiebib} 

\end{document}